\documentclass[10pt,twocolumn,letterpaper]{article}

\usepackage{styles}
\usepackage{times}
\usepackage{epsfig}
\usepackage{graphicx}
\usepackage{amsmath}
\usepackage{amssymb}
\usepackage{natbib}
\usepackage[thinc]{esdiff}
\usepackage{listings}

\usepackage{hyperref}
\hypersetup{
  colorlinks=true,
}

\newcommand*\samethanks[1][\value{footnote}]{\footnotemark[#1]}

\begin{document}

\title{Parametric Variational Linear Units (PVLUs) in Deep Convolutional Networks}
\author{
Shikhar Ahuja\thanks{Both authors contributed equally to this research.}\\
Montgomery High School\\
Skillman, NJ\\
{\tt\small ahujashikhar314@gmail.com}
\and
Aarush Gupta\samethanks\\
Montgomery High School\\
Skillman, NJ\\
{\tt\small aarush.g75@gmail.com}
}

\maketitle

\begin{abstract}
 The Rectified Linear Unit is currently a state-of-the-art activation function in deep convolutional neural networks. To combat ReLU's dying neuron problem, we propose the Parametric Variational Linear Unit (PVLU), which adds a sinusoidal function with trainable coefficients to ReLU. Along with introducing nonlinearity and non-zero gradients across the entire real domain, PVLU acts as a mechanism of fine-tuning when implemented in the context of transfer learning. On a simple, non-transfer sequential CNN, PVLU substitution allowed for relative error decreases of 16.3\% and 11.3\% (without and with data augmentation) on CIFAR-100. PVLU is also tested on transfer learning models. The VGG-16 and VGG-19 models experience relative error reductions of 9.5\% and 10.7\% on CIFAR-10, respectively, after the substitution of ReLU with PVLU. When training on Gaussian-filtered CIFAR-10 images, similar improvements are noted for the VGG models. Most notably, fine-tuning using PVLU allows for relative error reductions up to and exceeding 10\% for near state-of-the-art residual neural network architectures on the CIFAR datasets.  
\end{abstract}

\section{Introduction}

The Rectified Linear Unit, known as ReLU, has become a standard function in deep neural networks \citep{nair10relu}. Sigmoidal activation functions saturate as inputs deviate from zero, meaning that their gradients vanish and significant training fails to occur. This follows from noting that $\sigma'(x) = \sigma(x)(1 - \sigma(x))$ and $\sigma(x) \to 1$ as $x \to \infty$ or $x \to -\infty$.

However, ReLU suffers from its own key issue known as the dying neuron problem. Due to its derivative being zero for negative inputs, the use of ReLU may sometimes zero gradients and completely stop local weight updates during training \citep{gu2017recent}. The issue becomes even more prevalent for deeper networks — just a few dying neurons can zero the long chain rule expressions used to compute training gradients. It has been mathematically proven that the use of ReLU will necessarily lead to the dying neuron problem in the limit of an infinitely deep network \citep{Lu_2020}. 

Variants of ReLU have been created in the past to address its various issues. Three popular variants are the Exponential Linear Unit (ELU) \citep{clevert2016fast}, the Leaky Rectified Linear Unit (Leaky ReLU) \citep{maas13}, and the Parametric Rectified Linear Unit (PReLU) \citep{he2015delving}. All three have gotten results in a variety of datasets but have their respective issues. ELU's gradients saturate while Leaky ReLU and PReLU remain unbounded for negative domains.

\begin{figure}[h!]
\begin{center}
    \includegraphics[width=0.8\linewidth]{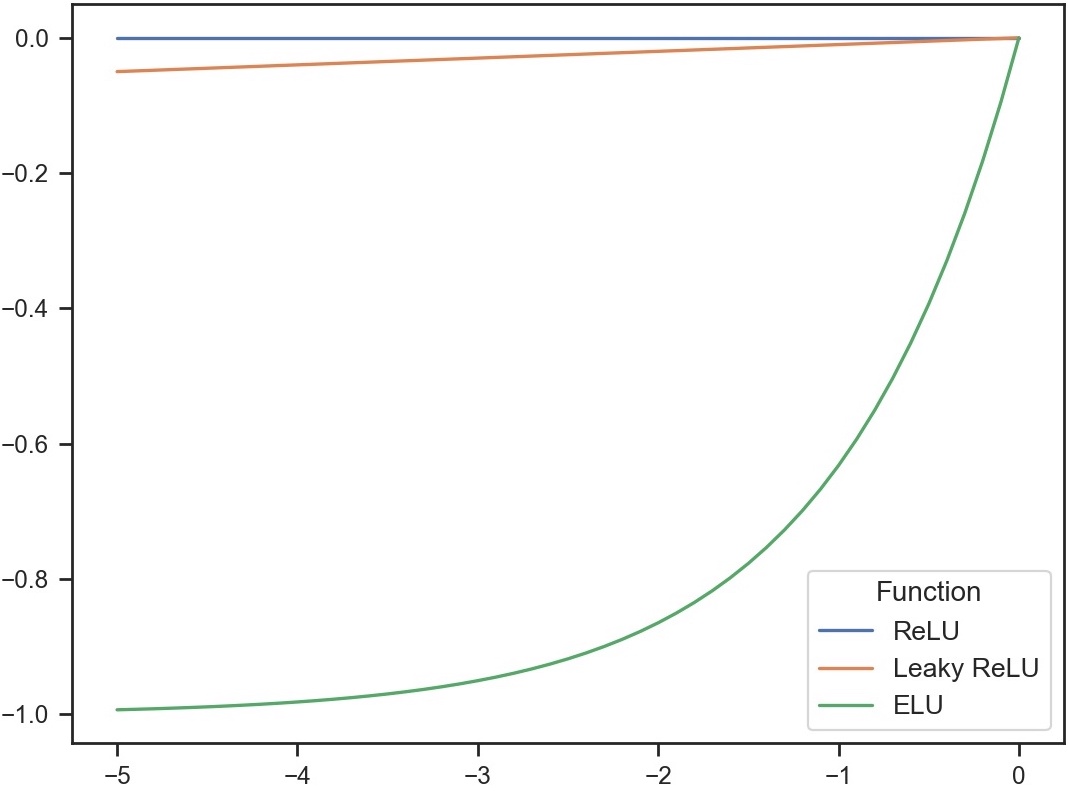}
\end{center}
\caption{Behavior of ReLU, Leaky ReLU, and ELU for $x < 0$.}
\label{fig:reluvariants}
\end{figure}

While less common, sinusoidal aspects have been incorporated in the past into activation functions \citep{sitzmann2020implicit}. In the context of our work, one relevant example is the SineReLU, available in Keras's contrib module \citep{chollet2015keras}. The function is written as the following (assuming $\epsilon$ is a hyperparameter):
\begin{equation*}
    \mathrm{SineReLU}(x) = \begin{cases} 
      x & \text{if $x\leq 0$}\\
      \epsilon(\sin{x} - \cos{x}) & \text{if $x > 1$}  
   \end{cases}
\end{equation*}

The SineReLU activation can assist with the dying neuron problem and we further extend on its findings. In doing so, we propose the Parametric Variational Linear Unit (PVLU), which is built up from first considering a simpler Variational Linear Unit (VLU). While it shares similarities to the SineReLU activation discussed before, PVLU introduces a variety of novel features that will be discussed further in our paper. 
In our paper, we first outline the VLU's theoretical basis and definition. Then, we extend our proposal to a parametrized VLU, or PVLU — applications in fine tuning and transfer learning are also discussed. Empirically, we demonstrate that PVLU performs better than ReLU on a variety of models running on the CIFAR-10 and CIFAR-100 dataset. These include everything from sequential, non-transfer CNNs to deep resiudal network models.

\section{Variational Linear Unit}

We first propose the Variational Linear Unit (VLU), which is defined as
\begin{align*}
    \mathrm{VLU}(x) &= \mathrm{ReLU}(x) + \alpha\sin{(\beta x)} \\
    &= \max(0, x) + \alpha\sin{(\beta x)}
\end{align*}
where $\alpha$ and $\beta$ are hyperparameters that can be chosen based on the model. Larger values for both of these parameters inject more periodic variation into the ReLU function. For $\alpha = 0$ or $\beta = 0$, no variation is present. 

Note that there are significant differences between the previously seen SineReLU and the currently proposed VLU. VLU introduces a sinusoidal variation throughout the entire real domain $x \in \mathbb{R}$. This differs from not only SineReLU, but also a significant number of previously seen ReLU variants which only alter outputs for domain $x < 0$. In doing so, we add nonlinearity throughout the entire activation function while also preserving one of ReLU's most important properties, which is its unbounded nature for positive values. Similar to SineReLU, adding non-monotonicity to the ReLU activation function may be beneficial as was observed with the Swish activation function \citep{ramachandran2017searching}.

The graphs below illustrate VLU's behavior for selected values of $\alpha$ and $\beta$.

\begin{figure}[h!]
\begin{center}
    \includegraphics[width=0.8\linewidth]{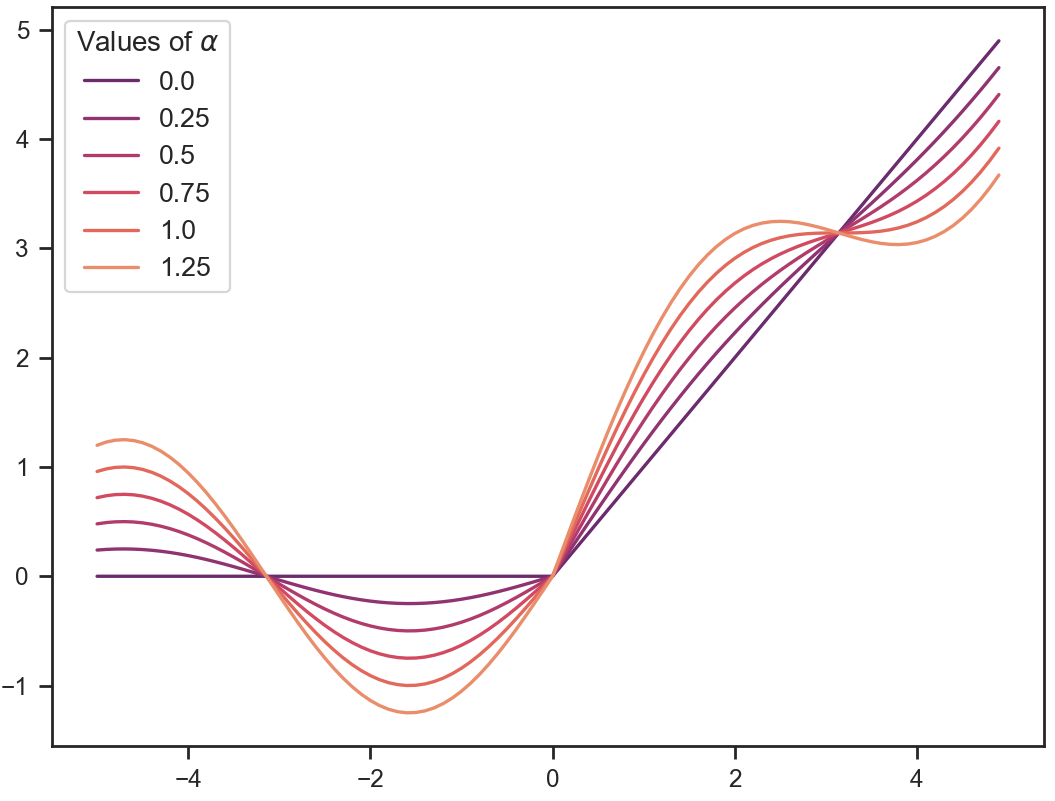}
\end{center}
\caption{Different values of $\alpha$ in $f(x) = \mathrm{ReLU}(x) + \alpha\sin{(\beta x)}$.}
\label{fig:alphavals}
\end{figure}

\begin{figure}[h!]
\begin{center}
   \includegraphics[width=0.8\linewidth]{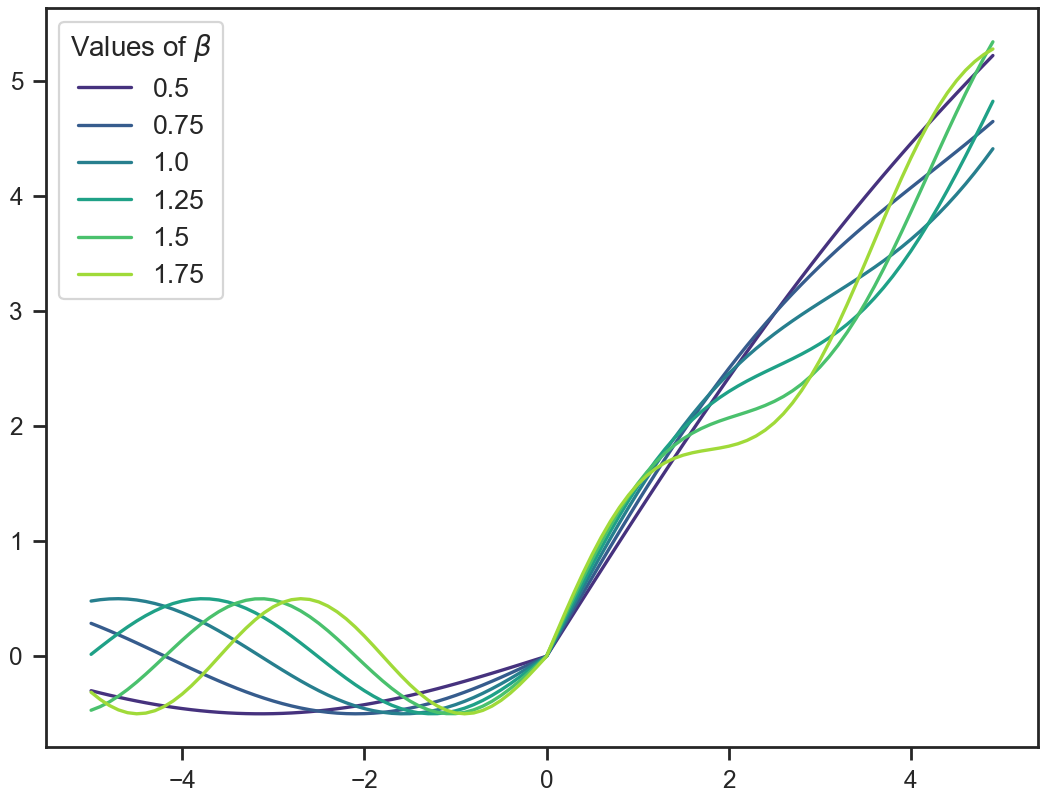}
\end{center}
\caption{Different values of $\beta$ in $f(x) = \mathrm{ReLU}(x) + \alpha\sin{(\beta x)}$.}
\label{fig:betavals}
\end{figure}

We can also analyze VLU's properties more rigorously. Let $\mathbf{a^\ell} \in \mathbb{R}^{N_\ell}$ be the set of activations $a_1, a_2, a_3, \dots, a_{N^{\ell}}$ in layer $\ell$, where $N_\ell$ is the number of neurons in that layer. Then, we define $z^\ell \in \mathbb{R}^{N_\ell}$ as $W^{\ell - 1}\bold{a}^{\ell - 1} + \bold{b}^{\ell - 1}$, where  $\bold{b^{\ell}} \in \mathbb{R}^{N_\ell}$ is the bias vector. $W^{\ell - 1} \in \mathbb{R}^{N_{\ell - 1} \times N_{\ell}}$ is the weight matrix connecting layers $\ell - 1$ and $\ell$.

Assuming weight $W_{ij}^{\ell}$ connects the $j$-th neuron in layer $\ell$ and the $i$-th neuron in layer $\ell + 1$, recall that
\begin{align*}
    z_i^{\ell + 1} &= \sum_{j = 1}^{N_{\ell}}\left(W_{ij}^{\ell}a_j^\ell + b_j^{\ell}\right) \\
    a_i^{\ell + 1} &= \alpha\sin{(\beta z_i^{\ell + 1})} + \mathrm{ReLU}( z_i^{\ell + 1})
\end{align*}
Then, the partial derivatives of the cost function $\mathcal{C}$ with respect to the network weights and biases are:
\begin{align*}
    \diffp{\mathcal{C}}{{{W_{ij}^\ell}}} &= \diffp{\mathcal{C}}{{{a_i^{\ell + 1}}}} \diffp{{{a_i^{\ell + 1}}}}{{{W_{ij}^\ell}}} \\ &= \diffp{\mathcal{C}}{{{a_i^\ell}}} a_j^{\ell}\left(\alpha\beta\cos{(\beta  z_i^{\ell + 1}) + H(z_i^{\ell + 1})}\right)\\
    \diffp{\mathcal{C}}{{{b_i^\ell}}} &= \diffp{\mathcal{C}}{{{a_i^\ell}}} \diffp{{{a_i^\ell}}}{{{b_i^\ell}}} \\ &= \diffp{\mathcal{C}}{{{a_i^\ell}}} \left(\alpha\beta\cos{(\beta  z_i^{\ell + 1}) + H(z_i^{\ell + 1})}\right)
\end{align*}
where 
\begin{equation*}
    H(x) = \begin{cases} 
      0 & \text{if $x\leq 0$}\\
      1 & \text{if $x > 0$}  
   \end{cases}
\end{equation*}
Note that these gradients are similar to that of ReLU except for the addition of the trigonometric terms. Since sine and cosine do not go to zero for large negative inputs, they help to alleviate the dying neuron problem present in ReLU. However, unlike ReLU, the derivative of our new activation function is no longer static for positive inputs.

\section{Parametric VLU}

Now we extend the discussion of the VLU to the Parametric VLU, or PVLU. Rather than static constants across the entire network, the parameters $\alpha$ and $\beta$ are now channelwise and are trained during backpropogation.

Then, our new activation function $A^\ell: \mathbb{R}^{N_\ell} \rightarrow \mathbb{R}^{N_\ell}$ in layer $\ell$ is defined by the channel-wise parameters $\alpha$ and $\beta$ like so:

\begin{equation*}
    A^\ell(\bold{z^\ell}) = \dottedcolumn{5}{\;\;\;\alpha_1^\ell \sin{(\beta_1^\ell  z_1^\ell) + \mathrm{ReLU}(z_1^\ell)}}{\alpha_{N^\ell}^\ell \sin{(\beta_{N^\ell}^\ell  z_{N^\ell}^\ell) + \mathrm{ReLU}(z_N^\ell)}} 
\end{equation*}
Using the chain rule, we can thus compute the gradients of the cost function $\mathcal{C}$.

\begin{align*}
    \diffp{\mathcal{C}}{{{\alpha_i^\ell}}} &= \diffp{\mathcal{C}}{{{a_i^\ell}}} \diffp{{{a_i^\ell}}}{{{\alpha_i^\ell}}} = \diffp{\mathcal{C}}{{{a_i^\ell}}} \sin{(\beta_i^\ell  z_i^\ell)}\\
    \diffp{\mathcal{C}}{{{\beta_i^\ell}}} &= \diffp{\mathcal{C}}{{{a_i^\ell}}} \diffp{{{a_i^\ell}}}{{{\beta_i^\ell}}} = \diffp{\mathcal{C}}{{{a_i^\ell}}}z_i^\ell\cos{(\beta_i^\ell  z_i^\ell)}
\end{align*}
 
As for the partial derivatives of the cost function with respect to weights and biases, we simply replace $\alpha$ and $\beta$ from our original expressions with their respective channel-wise parameters.
\begin{align*}
    \diffp{\mathcal{C}}{{{W_{ij}^\ell}}} &= \diffp{\mathcal{C}}{{{a_i^{\ell + 1}}}} \diffp{{{a_i^{\ell + 1}}}}{{{W_{ij}^\ell}}} \\ &= \diffp{\mathcal{C}}{{{a_i^\ell}}} a_j^{\ell}\left(\alpha_i^{\ell + 1}\beta_i^{\ell + 1}\cos{(\beta_i^{\ell+1}z_i^{\ell + 1}) + H(z_i^{\ell + 1})}\right)\\
    \diffp{\mathcal{C}}{{{b_i^\ell}}} &= \diffp{\mathcal{C}}{{{a_i^\ell}}} \diffp{{{a_i^\ell}}}{{{\beta_i^\ell}}} \\ &= \diffp{\mathcal{C}}{{{a_i^\ell}}} \left(\alpha_i^{\ell+1}\beta_i^{\ell+1}\cos{(\beta_i^{\ell+1} z_i^{\ell + 1}) + H(z_i^{\ell + 1})}\right)
\end{align*}

\subsection {Fine Tuning}
One of the major applications of PVLU in transfer learning is fine tuning. Given a pretrained deep model, all occurrences of ReLU can be replaced with PVLU where $\alpha$ is initialized to zero. Similar work with the already existing PReLU activation function (and even linear activations) has already been done in the past \citep{dong2017eraserelu}. This application allows PVLU to enhance models which are already performing at state-of-the-art accuracies.

Specifically, the use of PVLU can be thought of as adding a layer of trainable periodic noise to a network when applied in the context of transfer learning. Noisy activation functions have been shown to improve model performance \citep{gulcehre2016noisy}. What seperates this proposal from previous literature relating to noisy activation functions is that the added noise is not purely entropic. 

\section {Empirical Results}

Initially, we tested shallow architectures on various simple datasets such as MNIST and Fashion-MNIST. To test our proposal, we interchanged ReLU activations with PVLU. However, there was no significant difference or improvement after making this change. This result is logical as the dying neuron problem is not evident or impactful for networks that are not significantly deep. Thus, PVLU may add unnecessary complexity without real results for simple computer vision problems — the function's real applications are present in deeper and more complex networks.

\subsection {Initial Tests on CIFAR-100}

For the CIFAR-100 dataset, we utilized a deeper architecture where the dying neuron problem would be prevalent.  

For our first tests, we did not use data augmentation in assessing performance performance. To initially investigate the plausibility of PVLU, we test it on a relatively simple convolutional neural network. Our model is adapted from 
\begin{center}
\url{github.com/andrewkruger/cifar100_CNN}.
\end{center}

The model has six convolutional layers, each followed by a ReLU activation layer. The model also includes maxpooling, dropout, and dense layers. As a starting point, we run CIFAR-100 without data augmentation. In addition to testing ReLU as a baseline along with PVLU, we compare our proposal to a variety of existing variants (Leaky ReLU, ELU, PreLU). For the sake of consistency and reproducibility, we set random seeds using the Tensorflow function \lstinline{tf.random.set_seed} during each of our trials. As a starting point, the values of $\alpha$ and $\beta$ are set to $0.5$ and $1$, respectively. This ensures that the activations maintain a notable magnitude of sinusoidal nonlinearity.

Our peak test accuracies for five trials conducted are shown in the table below.

\begin{table}[h!]
\begin{center}
\begin{tabular}{c c c c c c} 
 Seed & ReLU & Leaky & PReLU & PVLU (ours) \\ [0.5ex] 
 \hline
 0 & 0.7\% & 50.1\% & 56.0\% & 56.6\%\\ 
 \hline
 1 & 49.7\% & 50.5\% & 55.5\% & 56.6\% \\
 \hline
 2 & 0.8\% & 50.5\% & 55.6\% & 57.8\%\\
 \hline
 3 & 48.7\% & 50.7\% & 54.9\% & 56.7\%\\
 \hline
 4 & 47.2\% & 50.3\% & 55.0\% & 56.7\%\\ 
 \hline\hline
 \textbf{Mean} & \textbf{48.5\%} & \textbf{50.4\%} & \textbf{55.5\%} & \textbf{56.9\%}\\
 \hline
 \textbf{Std. Err.} & \textbf{0.7\%} & \textbf{0.1\%} & \textbf{0.2\%} & \textbf{0.2\%}
\end{tabular}
\end{center}
\caption{Peak test accuracies of ReLU, PVLU, Leaky ReLU, and PReLU on simple non-transfer CNN without data augmentation.}
\end{table}
Note: ReLU "dies" several times in our trials. To fairly compare the activation functions, we only consider trials in which ReLU performs reasonably.

Examining the mean peak test accuracies of all activation functions along with their respective standard errors, PVLU performs the best with PReLU more than a percent in accuracy behind. These results can be clearly interpreted as statistically significant from comparing the standard errors in accuracies to their differences. As can be seen from the data, ReLU unexpectedly leads to a dying model for some trials — this may be reflective of its volatile nature as it relates to model initialization.

We can also examine the curves which show test accuracy as a function of epochs for each of the different activations. These curves are generated from the best trials of each function. 

\begin{figure}[h!]
\begin{center}
   \includegraphics[width=0.8\linewidth]{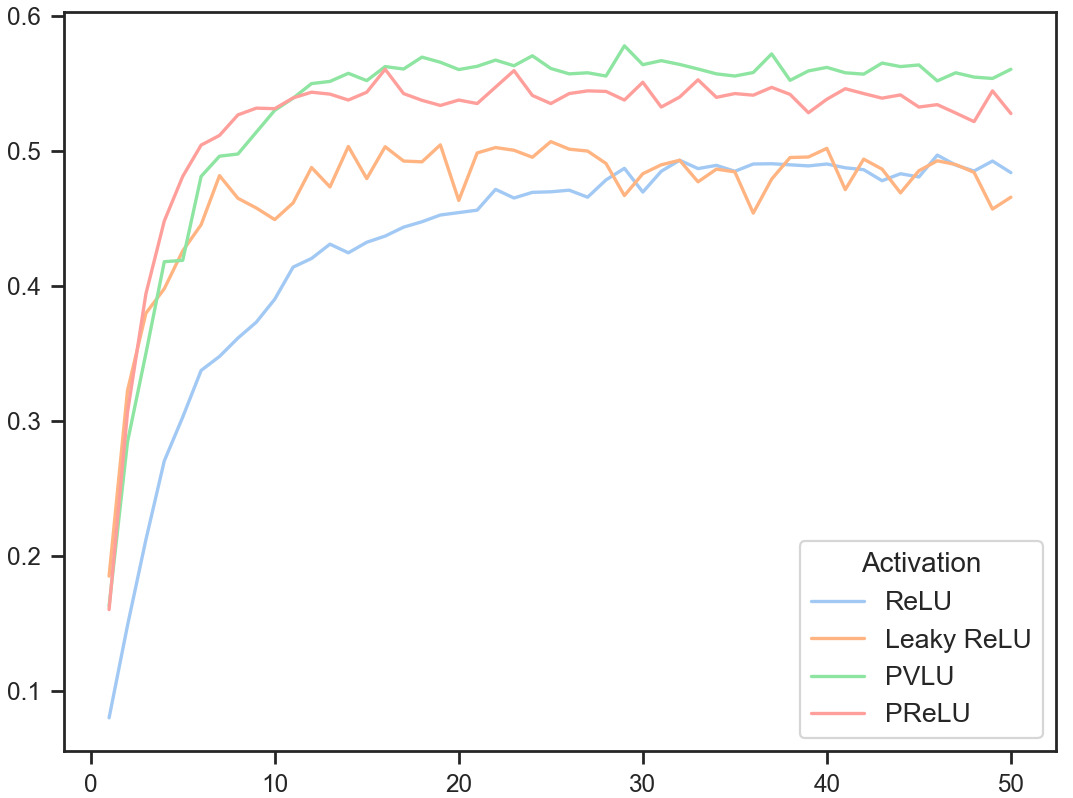}
\end{center}
   \caption{Test accuracy vs. epochs trained for ReLU, Leaky ReLU, PReLU, and PVLU on a non-transfer CNN without data augmentation.}
\label{fig:simpleaccs}
\end{figure}

PVLU and PReLU can be seen to converge far quicker than ReLU and its other variants. This fact can most likely be attributed to their parametrization and thus faster training.

While the above accuracies are expected for a relatively simple model, we also test PVLU's effects on more performant models. To do this, data augmentation is now applied while training on the same model. We use three trials instead of five due to computational limits. Additionally, Leaky ReLU is not included in our testing due to its subpar performance as compared to PReLU in previous tests. Our results are summarized below.

\begin{table}[h!]
\begin{center}
\begin{tabular}{c c c c c c} 
 Seed & ReLU & PReLU & PVLU (ours) \\ [0.5ex] 
 \hline
 0 & 61.9\% & 67.2\% & 67.4\%\\ 
 \hline
 1 & 61.9\% & 66.6\% & 65.9\%\\
 \hline
 2 & 61.7\% & 66.9\% & 64.9\%\\
 \hline\hline
 \textbf{Mean} & \textbf{61.8\%} & \textbf{66.9\%} & \textbf{66.1\%} \\
 \hline
 \textbf{Std. Error} & \textbf{0.1\%} & \textbf{0.7\%} & \textbf{0.1\%}
\end{tabular}
\end{center}
\caption{Peak test accuracies of ReLU, Leaky ReLU, PReLU, and PVLU (ours) on simple non-transfer CNN without data augmentation.}
\end{table}

With data augmentation, ReLU converged at 62 percent while PVLU and PReLU converged at 67\% — this model performance is on the higher side for a sequential model that does not employ transfer learning. Based on the standard error intervals, PVLU and PReLU performances were statistically similar. However, PVLU performed better when comparing top trials while PReLU had a higher mean accuracy.

Examining the trend of test accuracies for the best trial of each of the activation functions, it is seen that both PVLU and PReLU converge slower than ReLU (unlike our tests without data augmentation). Even approaching the hundredth epoch, PVLU and PReLU experience slow improvements in test accuracies.

\begin{figure}[h!]
\begin{center}
    \includegraphics[width=0.8\linewidth]{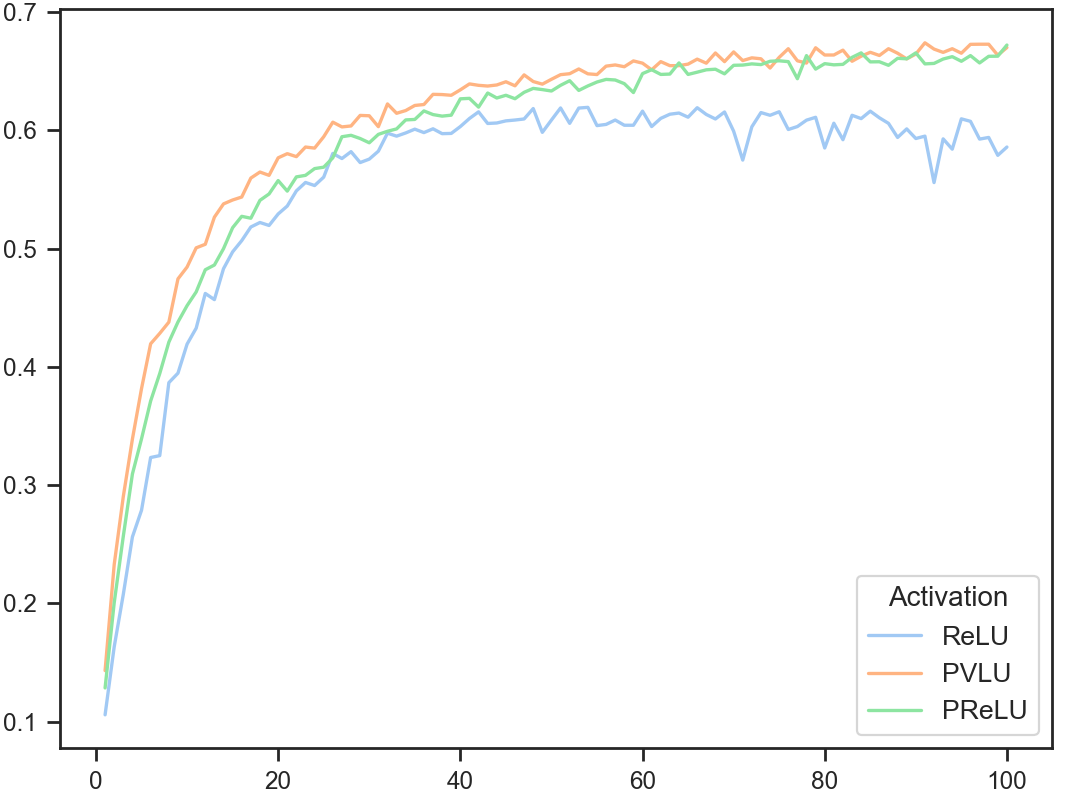}
\end{center}
   \caption{Test accuracy vs. epochs trained for ReLU, Leaky ReLU, PReLU, and PVLU on a non-transfer CNN with data augmentation.}
\label{fig:simpleaugaccs}
\end{figure}

\subsection {VGG-16 and VGG-19 Models}
VGG models are a powerful class of sequential convolutional networks that have achieved notable results on ImageNet and other deep computer vision problems \citep{simonyan2015deep}. With over one hundred million parameters, VGG models are perfect for transfer learning since complex representations are already prebuilt.

To test the performance of PVLU fine tuning, we first took the existing VGG models and trained their final layers on CIFAR-10 data. After obtaining the final convergence accuracies, we froze all model weights and replaced all instances of ReLU with PVLU ($\alpha = 0$, $\beta = 1$). This ensures that all improvements are solely the result of PVLU fine tuning. Our results are summarized in the table below.

\begin{table}[h!]
\begin{center}
\begin{tabular}{c c c c } 
 Model & ReLU & PVLU (ours) & Rel. Error Dec. \\ [0.5ex] 
 \hline
 VGG-16 & 86.95\% & 88.19\% & \textbf{9.5\%} \\ 
 \hline
 VGG-19 & 89.98\% & 91.06\% & \textbf{10.7\%} \\
\end{tabular}
\end{center}
\caption{VGG peak test accuracies for ReLU and PVLU on CIFAR-10. Relative error decrease is defined as $(e_i - e_f) / e_i$.}
\end{table}

The significant relative error reductions observed for both the VGG-16 and VGG-19 models highlight the effectiveness of PVLU fine tuning in transfer learning problems. 

\subsubsection{Noise Adaptability}

An important feature of a robust model is an ability to adapt to noise. To test PVLU in this context, a Gaussian filter is added to both training and testing CIFAR-10 data. Gaussian filters are used to blur images by performing convolution operations with a filter whose kernel coefficients follow the form
\begin{equation*}
   G(x, y) = \dfrac{1}{2\pi\sigma^2}\exp{\left(-\frac{x^2 + y^2}{\sigma^2}\right)} 
\end{equation*}
where $x$ and $y$ are the horizontal and vertical distances from the point at which the convolution is being performed. We set $\sigma = 1$ when applying the Gaussian filter to CIFAR-10 image data.

\begin{figure}[h!]
\begin{center}
    \includegraphics[width=0.45\linewidth]{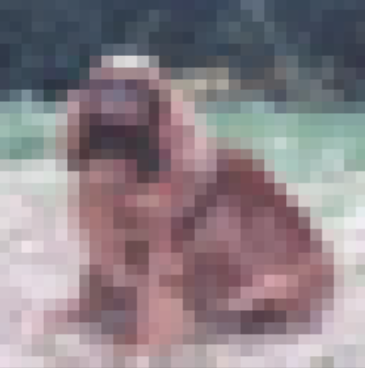}
    \includegraphics[width=0.45\linewidth]{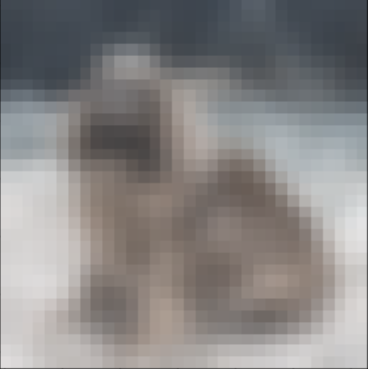}
\end{center}
\caption{CIFAR image before and after Gaussian filter is applied}
\label{fig:gaussian}
\end{figure}

First, we fine tune the existing VGG-19 model with ReLU activations on CIFAR-10. Then, this same model is evaluated on the noisy data to find a baseline accuracy. When testing PVLU, we freeze existing model weights while keeping the channelwise values of $\alpha$ and $\beta$ as trainable.
Allowing all layers to train in a noisy setting leads to sharp gradient updates that can interfere with and destroy representations in the existing transferred model. The only exception to this was for batch normalization layers — setting these to trainable allows for robust model generalization and adaptability \citep{ioffe2015batch}.

\begin{table}[h!]
\begin{center}
\begin{tabular}{c c c c } 
 Model & ReLU & PVLU (ours) & Rel. Error Dec. \\ [0.5ex] 
 \hline
 VGG-16 & 84.89\% & 86.30\% & \textbf{9.3\%} \\ 
 \hline
 VGG-19 & 82.34\% & 84.05\% & \textbf{9.7\%}\\
\end{tabular}
\end{center}
\caption{VGG peak test accuracies for ReLU and PVLU on CIFAR-10 noised using a Gaussian filter. Relative error decrease is defined as $(e_i - e_f) / e_i$.}
\end{table}

From the magnitude of relative error reductions observed, it is clear that PVLU can be used to adapt already trained and accurate models to new, noisy conditions. This approach, along with batch normalization, is far more practical than attempting to train an entirely new model on a noisy dataset.

\subsection {Residual Neural Networks}
In nearly all current state-of-the-art computer vision models, residual neural network architectures are used. Resiudal neural networks, or ResNets, use skip connections between non-adjacent layers to improving network training \citep{he2015deep}. In our tests, we fine-tune a pretrained ResNet-50 model on CIFAR-100, keeping only the batch normalization layers as trainable. We also test the ResNet-101 v2 architecture and ResNet-152 v2 architecture, which use identity mappings as residual connections \citep{he2016identity}. To ensure quick and manageable processing times, the original models containing ReLU were trained for 15 epochs. PVLU was then substituted and trained for another 15 epochs. When testing on ResNets, we use the cutout data augmentation technique as a method of regularization for our networks \citep{devries2017improved}.

\begin{table}[h!]
\begin{center}
\begin{tabular}{c c c c} 
 Model & ReLU & PVLU (ours) & Rel. Error Dec. \\ [0.5ex] 
 \hline
 R50 & 95.46\% & 95.82\% & \textbf{7.9}\%\\ 
 \hline
 R101v2 & 95.71\% & 96.20\% & \textbf{11.2}\%\\ 
 \hline 
 R152V2 & 96.39\% & 96.49\% & \textbf{2.8}\%\\
\end{tabular}
\end{center}
\caption{ResNet peak test accuracies on CIFAR-10. Relative error decrease is defined as $(e_i - e_f) / e_i$.}
\end{table}

\begin{table}[h!]
\begin{center}
\begin{tabular}{c c c c c} 
 Model & ReLU & PVLU (ours) & Rel. Error Dec. \\ [0.5ex] 
 \hline
 R50 & 79.9\% & 80.7\% & \textbf{4.1}\%\\
 \hline
 R101v2 & 80.66\% & 81.29\% & \textbf{3.3}\%\\
 \hline 
 R152v2 & 81.66\% & 82.52\% & \textbf{4.7}\%\\
\end{tabular}
\end{center}
\caption{ResNet peak test accuracies on CIFAR-100. Relative error decrease is defined as $(e_i - e_f) / e_i$.}
\end{table}

The results are especially significant because they demontrate PVLU's capacity for improvement on models which achieve near state-of-the-art results. It is worth noting that the original time of training (15 epochs) was enough time for the models to converge or come extremely close to converging, because all layers were frozen except for batch normalization. In the tests that we ran, the relative error decreases on ResNet models indicates that PVLU has significant potential in transfer models.

\section{Conclusion}
The Parametric Variational Linear Unit, or PVLU, leads to notable improvements across a wide variety of computer vision problems. By introducing a sinusoidal nonlinearity across the ReLU function, PVLU diminishes the dying neuron problem and also increases activation nonlinearity. In our empirical work, we tested PVLU across shallow non-transfer convolutional neural networks, VGG models, and residual neural networks. We also demonstrated that PVLU can lead to increased robustness when a previously-trained model is introduced to a noisy environment. Observed relative error reductions of roughly $10\%$ or more indicate that PVLU is worth further exploring in different contexts.

While this work primary centered around computer vision problems, PVLU may also be investigated in other domains of machine learning problems in the future. Within computer vision, PVLU can be tested on other transfer learning models for training deep networks on even more complex datasets.

{\small
\bibliographystyle{unsrtnat}
\bibliography{references}

\begin{thebibliography}{16}
\providecommand{\natexlab}[1]{#1}
\providecommand{\url}[1]{\texttt{#1}}
\expandafter\ifx\csname urlstyle\endcsname\relax
  \providecommand{\doi}[1]{doi: #1}\else
  \providecommand{\doi}{doi: \begingroup \urlstyle{rm}\Url}\fi

\bibitem[Nair and Hinton(2010)]{nair10relu}
Vinod Nair and Geoffrey~E. Hinton.
\newblock Rectified linear units improve restricted boltzmann machines.
\newblock In Johannes Fürnkranz and Thorsten Joachims, editors, \emph{ICML},
  pages 807--814. Omnipress, 2010.

\bibitem[Gu et~al.(2017)Gu, Wang, Kuen, Ma, Shahroudy, Shuai, Liu, Wang, Wang,
  Wang, Cai, and Chen]{gu2017recent}
Jiuxiang Gu, Zhenhua Wang, Jason Kuen, Lianyang Ma, Amir Shahroudy, Bing Shuai,
  Ting Liu, Xingxing Wang, Li~Wang, Gang Wang, Jianfei Cai, and Tsuhan Chen.
\newblock Recent advances in convolutional neural networks, 2017.

\bibitem[Lu et~al.(2020)Lu, Shin, Su, and Em~Karniadakis]{Lu_2020}
Lu~Lu, Yeonjong Shin, Yanhui Su, and George Em~Karniadakis.
\newblock Dying relu and initialization: Theory and numerical examples, 2020.
\newblock ISSN 1991-7120.

\bibitem[Clevert et~al.(2016)Clevert, Unterthiner, and
  Hochreiter]{clevert2016fast}
Djork-Arné Clevert, Thomas Unterthiner, and Sepp Hochreiter.
\newblock Fast and accurate deep network learning by exponential linear units
  (elus), 2016.

\bibitem[Maas et~al.(2013)Maas, Hannun, and Ng]{maas13}
Andrew~L. Maas, Awni~Y. Hannun, and Andrew~Y. Ng.
\newblock Rectifier nonlinearities improve neural network acoustic models,
  2013.

\bibitem[He et~al.(2015{\natexlab{a}})He, Zhang, Ren, and Sun]{he2015delving}
Kaiming He, Xiangyu Zhang, Shaoqing Ren, and Jian Sun.
\newblock Delving deep into rectifiers: Surpassing human-level performance on
  imagenet classification, 2015{\natexlab{a}}.

\bibitem[Sitzmann et~al.(2020)Sitzmann, Martel, Bergman, Lindell, and
  Wetzstein]{sitzmann2020implicit}
Vincent Sitzmann, Julien N.~P. Martel, Alexander~W. Bergman, David~B. Lindell,
  and Gordon Wetzstein.
\newblock Implicit neural representations with periodic activation functions,
  2020.

\bibitem[Chollet et~al.(2015)]{chollet2015keras}
Francois Chollet et~al.
\newblock Keras, 2015.
\newblock URL \url{https://github.com/fchollet/keras}.

\bibitem[Ramachandran et~al.(2017)Ramachandran, Zoph, and
  Le]{ramachandran2017searching}
Prajit Ramachandran, Barret Zoph, and Quoc~V. Le.
\newblock Searching for activation functions, 2017.

\bibitem[Dong et~al.(2017)Dong, Kang, Zhan, and Yang]{dong2017eraserelu}
Xuanyi Dong, Guoliang Kang, Kun Zhan, and Yi~Yang.
\newblock Eraserelu: A simple way to ease the training of deep convolution
  neural networks, 2017.

\bibitem[Gulcehre et~al.(2016)Gulcehre, Moczulski, Denil, and
  Bengio]{gulcehre2016noisy}
Caglar Gulcehre, Marcin Moczulski, Misha Denil, and Yoshua Bengio.
\newblock Noisy activation functions, 2016.

\bibitem[Simonyan and Zisserman(2015)]{simonyan2015deep}
Karen Simonyan and Andrew Zisserman.
\newblock Very deep convolutional networks for large-scale image recognition,
  2015.

\bibitem[Ioffe and Szegedy(2015)]{ioffe2015batch}
Sergey Ioffe and Christian Szegedy.
\newblock Batch normalization: Accelerating deep network training by reducing
  internal covariate shift, 2015.

\bibitem[He et~al.(2015{\natexlab{b}})He, Zhang, Ren, and Sun]{he2015deep}
Kaiming He, Xiangyu Zhang, Shaoqing Ren, and Jian Sun.
\newblock Deep residual learning for image recognition, 2015{\natexlab{b}}.

\bibitem[He et~al.(2016)He, Zhang, Ren, and Sun]{he2016identity}
Kaiming He, Xiangyu Zhang, Shaoqing Ren, and Jian Sun.
\newblock Identity mappings in deep residual networks, 2016.

\bibitem[DeVries and Taylor(2017)]{devries2017improved}
Terrance DeVries and Graham~W. Taylor.
\newblock Improved regularization of convolutional neural networks with cutout,
  2017.

\end{thebibliography}
}

\end{document}